\documentclass[runningheads]{llncs}
\usepackage{graphicx}
\usepackage{hyperref}

\usepackage{algorithm}
\usepackage{algorithmic}

\usepackage{amsmath,amssymb}
\usepackage{multirow}
\usepackage{cleveref}
\usepackage{makecell}

\crefname{figure}{Fig.}{Fig.}
\crefname{equation}{Eq.}{Eq.}
\crefname{table}{Table}{Table}
\crefname{section}{Section}{Section}

\usepackage{xcolor}

\usepackage{mathtools}
\DeclarePairedDelimiter{\norm}{\lVert}{\rVert}

\newcommand{\paragraphbf}[1]{\noindent \textbf{#1}\hspace{1em}}

\usepackage[labelformat=simple]{subcaption}

\usepackage{xspace}
\makeatletter
\DeclareRobustCommand\onedot{\futurelet\@let@token\@onedot}
\def\@onedot{\ifx\@let@token.\else.\null\fi\xspace}
\def\eg{\emph{e.g}\onedot} 
\def\ie{\emph{i.e}\onedot}

\makeatother

\usepackage{enumitem}
\newlist{inparaenum}{enumerate}{2}%
\setlist[inparaenum]{nosep}%
\setlist[inparaenum,1]{label=\arabic*.}%
\setlist[inparaenum,2]{label=\arabic{inparaenumi}\emph{\alph*})}%

\begin{document}
\title{Adversarial Augmentation Training Makes Action Recognition Models More Robust to Realistic Video Distribution Shifts}
\titlerunning{Adversarial Augmentation Training for Action Recognition}
\author{Kiyoon Kim \and
Shreyank N Gowda \and
Panagiotis Eustratiadis \and
Antreas Antoniou \and
Robert B Fisher
}
\authorrunning{K. Kim et al.}
\institute{University of Edinburgh, Edinburgh, UK}
\maketitle              %
\begin{abstract}
Despite recent advances in video action recognition achieving strong performance on existing benchmarks, these models often lack robustness when faced with natural distribution shifts between training and test data. 
We propose two novel evaluation methods to assess model resilience to such distribution disparity.
One method uses two different datasets collected from different sources and uses one for training and validation, and the other for testing.
More precisely, we created dataset splits of HMDB-51 or UCF-101 for training, and Kinetics-400 for testing, using the subset of the classes that are overlapping in both train and test datasets.
The other proposed method extracts the feature mean of each class from the target evaluation dataset's training data ({\it i.e.} class prototype), and estimates test video prediction as a cosine similarity score between each sample to the class prototypes of each target class.
This procedure does not alter model weights using the target dataset and it does not require aligning overlapping classes of two different datasets, thus it is a very efficient method to test the model robustness to distribution shifts, without prior knowledge of the target distribution.
We address the robustness problem by adversarial augmentation training -- generating augmented views of videos that are ``hard'' for the classification model by applying gradient ascent on the augmentation parameters -- as well as ``curriculum'' scheduling the strength of the video augmentations. 
We experimentally demonstrate the superior performance of the proposed adversarial augmentation approach over baselines across three state-of-the-art action recognition models - TSM, Video Swin Transformer, and Uniformer.
The presented work provides critical insight into model robustness to distribution shifts and presents effective techniques to enhance video action recognition performance in a real-world deployment.

\end{abstract}
\section{Introduction}

\begin{figure}[t]
    \centering
    \includegraphics[width=0.7\columnwidth]{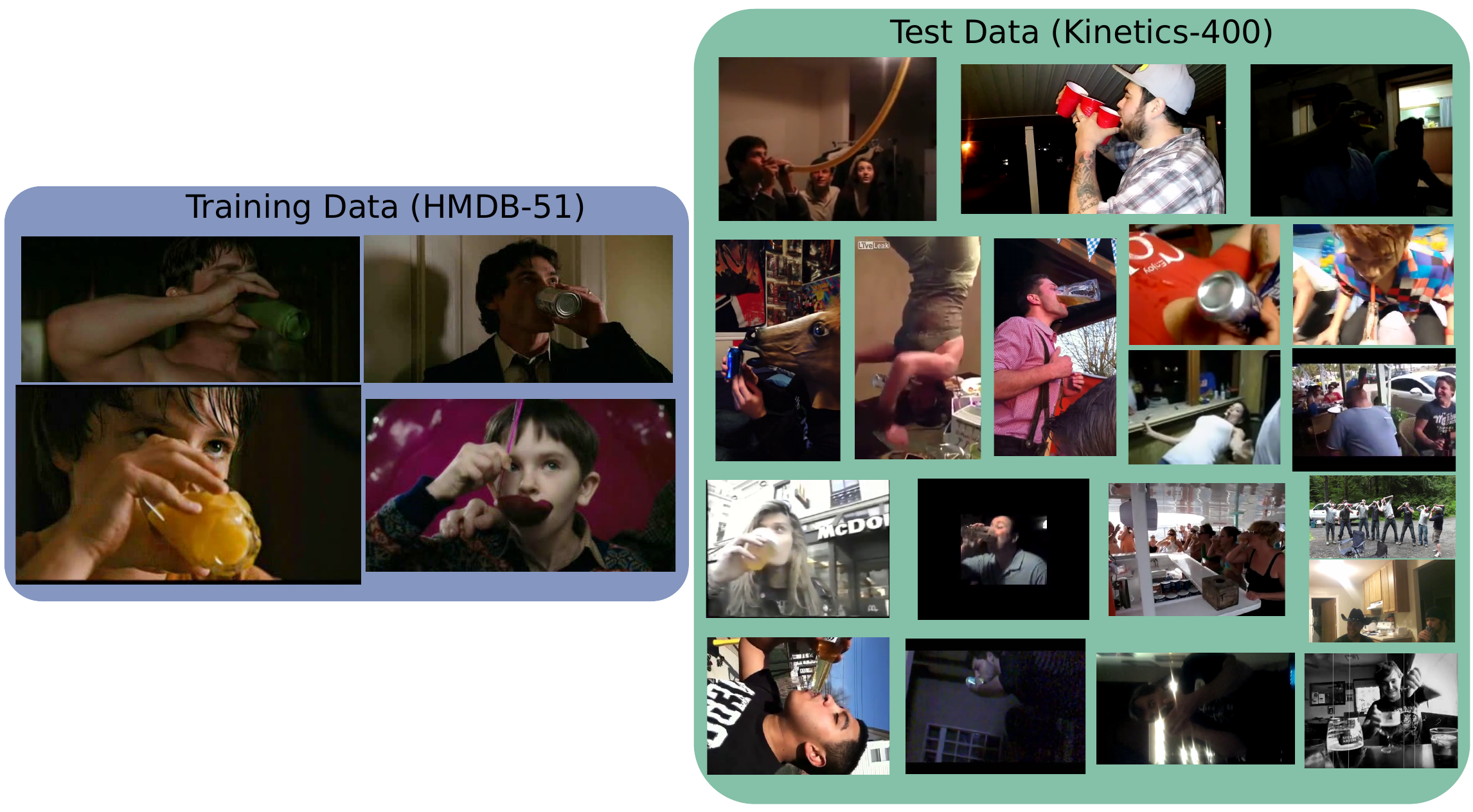}
    \vspace{-5pt}
    \caption{A common scenario with biased training data and testing with real-life data, with an example class of ``drink''. The training data is from the HMDB-51 dataset whose video samples are usually taken from movies, and the test data is from the Kinetics-400 dataset which is from YouTube videos. 
    There are many reasons why the test data looks so different: poor camera quality, wrong orientation, extreme camera shake, inconsistent frame rate, frame rate conversion artifacts (interlacing), poor lighting, lack of professional post-processing (\eg color grading), different ways of performing the action, poor framing, inconsistent aspect ratios, editing artifacts, various actions happening at the same time. 
    Thus, it is common for the performance to drop significantly when the trained model is applied in more general applications.}
    \label{fig:drink}
    \vspace{-1em}
\end{figure}

Video action recognition is a vital computer vision task with applications in surveillance, robotics, and more. 
Video data exhibits greater diversity than image data,
 and therefore action recognition architectures are not as robust to distribution shifts~\cite{yi2021benchmarking,Schiappa_2023_CVPR,lin2023video}. 
 In addition to image-level effects like viewpoint and appearance changes, video introduces effects such as camera motion, focus shifts, and background object movements. 
Moreover, an action class incorporates substantial intra-class variation as illustrated in \Cref{fig:drink}. For example, the class ``playing basketball" may involve dribbling, running, or shooting in different contexts. 
Furthermore, depending on the data source, there are biased video processing artifacts. 
For example, videos collected from YouTube have standardized YouTube processing (VP9 compression), making the dark areas have extremely low quality. 
Often, the videos go through a frame rate conversion algorithm, which can create frame interlacing artifacts. 
As a result, the slight distribution shift in video data can dramatically reduce the action recognition performance. 

Data augmentation is one potential solution to account for this fragility.
It is a popular method to create synthetic variations of the existing training data that will enable classification models to generalize better to previously unseen test data. 
However, it is not yet clear what kind of augmentation is necessary to generalize to test data with different distribution shifts. 
There has been much work on automatically selecting augmentation policies given training and validation data~\cite{cubuk2018autoaugment,ho2019population,fastautoaugment,fasterautoaugment,li2020differentiable}. 
However, such approaches optimize augmentation of the training data, and it is not clear how well the resulting generalization applies to test data with much distribution shift.

To address the video domain shift problem, we propose an adversarial augmentation scheme that generates ``hard'' video examples for the action recognition networks. 
The pipeline is simple to implement, and results in a meaningful improvement in performance on the proposed datasets with realistic distribution shifts compared to no augmentation and random augmentation baselines.
The benefits are demonstrated using three popular action recognition architectures. 
The training and validation datasets are subsets of HMDB-51 or UCF-101, and the test data are a subset from equivalent Kinetics-400 classes, to realistically evaluate distribution shifts over the same action classes.

This approach and evaluation requires multiple datasets with common aligned action classes. 
The paper also presents another simple method to evaluate robustness using a target dataset with different classes using cosine similarity of features as logits. 
The method requires no training (\ie fine-tuning) on the target dataset, and thus it correctly measures the transferability of the trained classifier on the target dataset.

To summarize, our contributions are:

\begin{inparaenum}
\item Experiments reveal substantial performance degradation on our cross-dataset benchmarks, quantitatively demonstrating the challenge posed by real-world distribution shift.
\item We propose two novel evaluation protocols to assess model resilience to distribution disparity using naturally sourced datasets, as opposed to solely artificially corrupted data:
\begin{inparaenum}
\item We construct new cross-dataset benchmarks by identifying overlapping classes between HMDB-51, UCF-101, and Kinetics-400. Models are trained on either HMDB or UCF, and evaluated on Kinetics data.

\item We further introduce a similarity-based evaluation approach that estimates predictions using cosine similarity between embedded training and test features, without requiring class alignment.
\end{inparaenum}
\item Through extensive experiments across multiple state-of-the-art architectures, we empirically demonstrate that the proposed adversarial augmentation and curriculum adversarial training frameworks enhance robustness to realistic distribution shifts between the training and test datasets.
\item We publicly release the constructed subsets of HMDB-51, UCF-101, and Kinetics-400, along with all experimental code, to enable further research on this important problem.\footnote{\url{https://github.com/kiyoon/video-adversarial-augmentation}}
\end{inparaenum}

\section{Related Work}

\paragraphbf{Action recognition.} Action recognition is the task of categorizing video sequences into predefined action classes. 
Architectures based on 3D convolutional neural networks were previously dominant for spatiotemporal feature learning. These include approaches such as inflating 2D models~\cite{i3d}, incorporating relational reasoning with non-local operations~\cite{nonlocal}, and dual-stream designs~\cite{twostream,slowfast}. 
More recently, transformer networks have become prominent, demonstrating strong performance~\cite{mvit,swin,li2022uniformer} despite their exponential computational complexity~\cite{kim2022capturing}. 
For efficient video recognition, 2D backbone models remain popular, using techniques like temporal feature aggregation~\cite{tsn}, relational modeling~\cite{trn}, temporal shift modules~\cite{tsm}, frame selection~\cite{scsampler,smart,adaframe}, channel-wise convolutions (\ie height-width, height-time, width-time)~\cite{wu2021mvfnet} or analyzing short-term and long-term temporal difference~\cite{tdn}.

\paragraphbf{Data augmentation.}
\cite{shorten2019survey} summarizes image data augmentation techniques for deep learning. AutoAugment~\cite{cubuk2018autoaugment} is an augmentation policy search algorithm that finds the best augmentation on a target dataset, based on the highest validation accuracy. 
Due to AutoAugment's expensive policy search, Population-Based Augmentation~\cite{ho2019population}, FastAutoAugment~\cite{fastautoaugment}, and FasterAutoAugment~\cite{fasterautoaugment} proposed more efficient searching algorithms, by learning a schedule policy over a fixed-policy, using density matching, and using differential augmentation with a generative adversarial network (GAN)~\cite{gan} architecture that involves a policy generator and a discriminator, respectively. 
Differentiable Automatic Data Augmentation~\cite{li2020differentiable} proposed a data augmentation policy searching algorithm (using the Relaxed Bernoulli distribution~\cite{jang2017categorical}) which is differentiable, similar to FasterAutoaugment, and further introduced an unbiased gradient estimator that enables joint optimization of the augmentation policies and network parameters, instead of using a GAN.
RandAugment~\cite{cubuk2020randaugment} showed that simple random augmentations with randomly sampled transformations achieve similar performance more efficiently.

However, the policies in most works are optimized on the training set, and we focus on the scenario where test data can have severe distribution disparities which are unknown during the training time.

\paragraphbf{Adversarial training.} 
Adversarial training (AT) is framed as a min-max problem whereby the trained model uses observed training samples to minimize its prediction error, while an adversary attempts to generate training samples that maximize it. 
It is well-established that AT is the most effective way to achieve adversarial robustness~\cite{ijcai21at}. It has further been shown to yield other types of robustness, \eg against natural corruptions~\cite{Dong_2023_CVPR}, domain shift~\cite{iclr18advdg,aaai20advdg,iclr23advdg}, and others. Note that even though the classical definition of AT uses adversarial input noise~\cite{iclr15fgsm,sp17cnw}, more adversarial image augmentations have been studied, \eg rotations~\cite{cvpr20isometry,cvpr22artpoint}, contrast, jitter, etc.~\cite{icbinb21imageaug}.
AT should be approached with care, as generating adversarial training examples that are too challenging for the trained model may actually harm downstream performance~\cite{cai2018curriculum}. 

In this paper, we employ two measures to control the trade-off between augmentation strength and performance:
(i) We create maximally informative adversarial examples (confusing to the model, but near the classification boundaries) via maximum-entropy regularization, as per the work of~\cite{cvpr20l2p,aaai21sesnn,icml21wca}. (ii) We train with curriculum AT as per the work of~\cite{cai2018curriculum,zhang2020attacks}, 
which means training with harder adversarial examples over time.

\paragraphbf{Domain adaptation.} Domain adaptation is a transfer learning task where the source and target datasets have a significant distribution shift while sharing the same task. 
\cite{farahani2021brief} explains types of domain adaptation tasks and approaches. 

There are discrepancy-based techniques that learn transferable features from a source domain to a target domain~\cite{long2015learning,zhang2015deep}, reconstruction-based methods that utilize autoencoders, which aim to extract useful features for the target domain~\cite{glorot2011domain,ghifary2016deep}, and adversarial domain adaptation approaches involving a source / target discriminator that distinguishes where the data come from and a feature extractor that aims to confuse the discriminator by trying to produce generic features regardless of the domain~\cite{ganin2015unsupervised,pei2018multi,tzeng2015simultaneous,ajakan2014domain,bousmalis2017unsupervised,shrivastava2017learning,taigman2016unsupervised,hoffman2018cycada}. 
More recently, analyzing frequency components of deep feature maps using attention to filter domain-general components~\cite{Lin_2023_CVPR} is proposed. 

Domain adaptation for video action recognition was first proposed by using a feature alignment approach on online test videos~\cite{lin2023video}. 
This work was evaluated using computationally simulated corrupted videos, while we propose to use real examples that involve more diverse types of discrepancy between the domains.

It is important to note that most domain adaptation techniques require examples from the target dataset to be present, while our work focuses on evaluating using a completely unknown dataset.

\paragraphbf{Corruption robustness analysis.} \cite{hendrycks2019robustness} provided benchmarks for measuring a neural network's robustness to corruptions and perturbations, by evaluating with 15 algorithmically-generated corruptions
(\eg noise, blur, pixelate, compression artifacts).
\cite{yi2021benchmarking} extended this to video classification tasks and video corruptions
(\eg video compression artifacts, frame rate conversion, bit error, packet loss). 
\cite{Schiappa_2023_CVPR} reported a large-scale robustness analysis of deep action recognition models again using pre-defined perturbations. 

These approaches were evaluated using simulated data, while we propose to use real data for testing. Evaluating robustness with augmented data prohibits the same augmentations to be used for training.
This paper focuses on a more realistic scenario where a known set of data augmentation strategies is used for training and evaluation is done with unprocessed real data.

\section{Problem and Methodology}

Action recognition predicts an action category label given a video sequence. 
This paper explores how well different variations of
action recognition models, training, and loss functions generalize by evaluating on a different data domain.
  
The main difference with transfer learning is that our approach does not tune model parameters using the target evaluation dataset, whereas transfer learning usually involves fine-tuning the model with the target dataset's training set.

To improve generalizability,  the training data is augmented. 
Hard-to-classify adversarial examples are generated by applying gradient ascent on the augmentation parameters which are fully differentiable. 
We then train the classifier using the AT (Adversarial Training) loss, calculated using both clean and adversarial examples.

\subsection{Adversarial Augmentation Training}
\label{sec:adversarial_method}
\begin{figure}[t!]
    \centering
    \includegraphics[width=0.7\columnwidth]{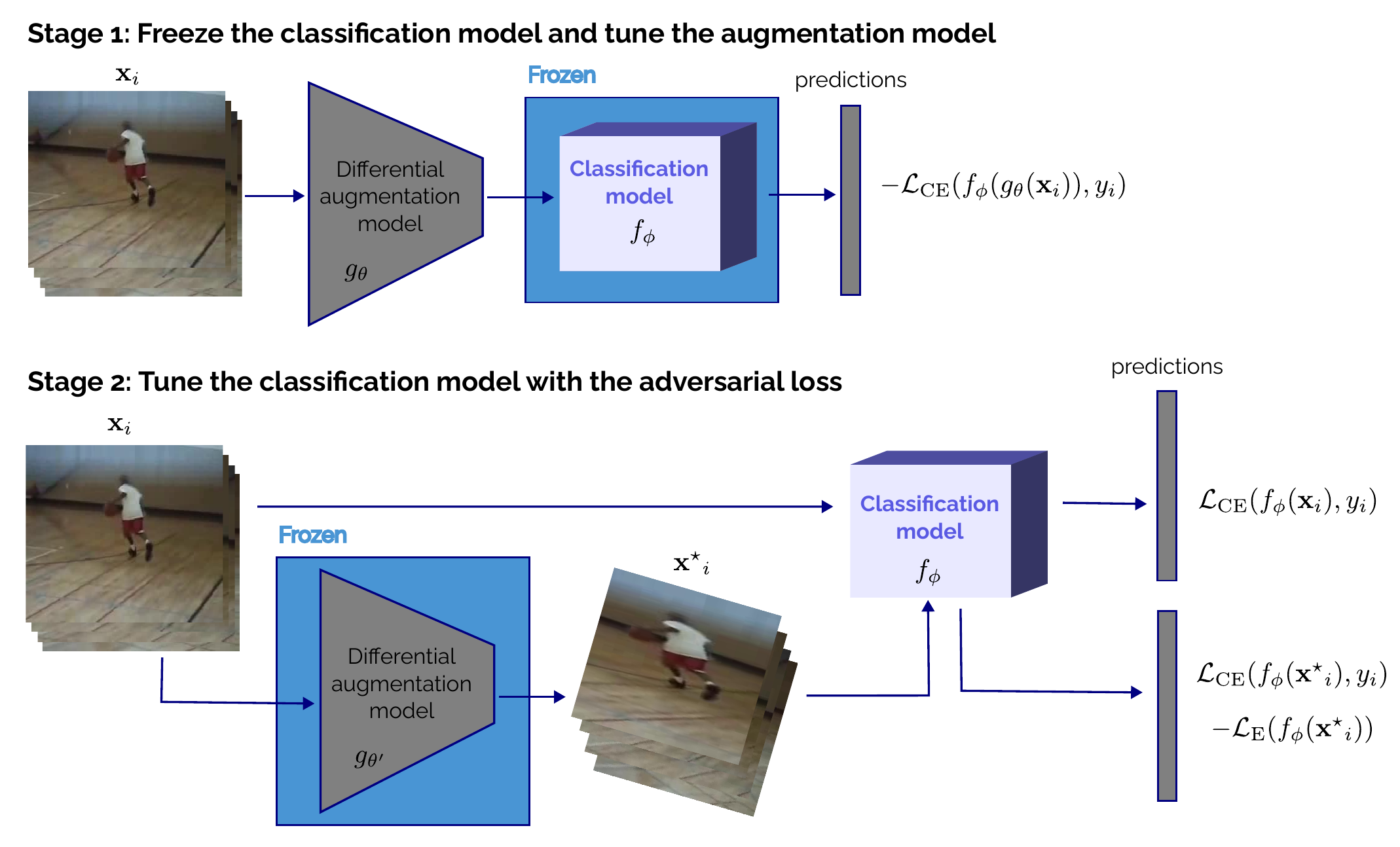}
    \vspace{-10pt}
    \caption{The proposed adversarial augmentation training has two separate stages. 
    Firstly, the classification model is frozen while the differential augmentation model is trained using the negative cross-entropy loss. 
    This is equivalent to performing gradient ascent to maximize normal cross-entropy loss. 
    As a result, the augmentation model will generate hard augmentations for the classification model. 
    The second stage trains the classification model using both clean and adversarial examples. 
    The maximum entropy regularization loss is integrated by subtracting the entropy of the adversarial examples, encouraging the predictions to be evenly balanced.}
    \label{fig:adversarial}
    \vspace{-1em}
\end{figure}

Adversarial augmentation training uses a two stage training loop. See \Cref{fig:adversarial}.

\paragraphbf{Stage 1: Generate adversarial examples.} ``Hard'' adversarial examples are found by tuning the augmentation parameters using gradient ascent.

Let $g_\theta(\mathbf{x})$ be an augmentation model with fully differentiable parameters $\theta$, and $f_\phi(\mathbf{x})$ be a video classification model with parameters $\phi$, that outputs class predictions given an input video $\mathbf{x}$. 
The goal of stage 1 is to find the augmentation parameters $\theta'$ that maximize categorical cross-entropy loss.
Note that by \emph{maximizing} the loss, we aim to find the augmentation strategy that is challenging for the classifier, and in \Cref{fig:adversarial}, this is described as \emph{minimizing} the negative cross-entropy loss.

At each training step,  the augmentation parameters $\theta$ are randomly initialized. 
Gradient ascent is then done only on $\theta$, freezing the classification parameters $\phi$ to learn adversarial augmentations, with the cross-entropy loss $\mathcal{L_{\text{CE}}}(f_\phi(g_\theta(\mathbf{x}_i)), y_i)$. 
This optimizes augmentation parameters $\theta'$ for generating adversarial examples.

\paragraphbf{Stage 2: Optimize the classification model.} Next, the classification parameters $\phi$ are optimized while freezing the augmentation parameters $\theta$. 
For simplicity,  $\mathbf{x}^\star_i = g_{\theta'}(\mathbf{x}_i)$ denotes the generated adversarial example of $\mathbf{x}_i$. 
The vanilla AT loss is defined as:
\begin{equation}
\label{equ:adversarial}
    \mathcal{L_{\text{AT}}}(f_\phi(\mathbf{x}_i), f_\phi(\mathbf{x}^\star_i), y_i) = \alpha\mathcal{L_{\text{CE}}}(f_\phi(\mathbf{x}_i), y_i) + (1-\alpha)\mathcal{L_{\text{CE}}}(f_\phi(\mathbf{x^\star}_i), y_i)
\end{equation}
\noindent which is a weighted average of the cross-entropy loss using the clean sample and the augmented sample.

\paragraphbf{Max-Entropy Regularization.} The cross-entropy loss encourages predictions to be over-confident by pushing the examples further from the classification boundaries. 
However, adversarial examples are supposed to be confusing. 
We regularize the cross-entropy-based adversarial loss in \Cref{equ:adversarial} by maximizing the entropy, encouraging the overall predictions to be evenly balanced for adversarial examples.
\begin{align}
\begin{split}
    \mathcal{L_{\text{AT-ME}}}(f_\phi(\mathbf{x}_i), f_\phi(\mathbf{x}^\star_i), y_i) &= \alpha\mathcal{L_{\text{CE}}}(f_\phi(\mathbf{x}_i), y_i) \\
    &+ (1-\alpha)\mathcal{L_{\text{CE}}}(f_\phi(\mathbf{x^\star}_i), y_i) - \gamma\mathcal{L_{\text{E}}}(f_\phi(\mathbf{x^\star}_i))
\end{split}
\end{align}
\noindent where entropy loss $\mathcal{L_{\text{E}}}$ is defined as
\begin{align}
    \mathcal{L_{\text{E}}}(f_\phi(\mathbf{x}^\star_i)) = -\frac{1}{C}\sum_{c=1}^{C}{f_\phi^{(c)}(\mathbf{x}^\star_i)\log(f_\phi^{(c)}(\mathbf{x}^\star_i))}
\end{align}
\noindent $C$ is the number of classes, and $f_\phi^{(c)}(\mathbf{x}^\star_i)$ is the prediction score of class $c$ for the adversarial example $\mathbf{x}^\star_i$.

\paragraphbf{Curriculum Adversarial Training.} Applying curriculum training by starting from training with ``easy'' samples and gradually generating ``harder'' samples  makes the model more robust~\cite{cai2018curriculum}. 
Here, the classification models are trained initially from clean data without augmentation, and gradually harder adversarial examples are added by scheduling the learning rate of the augmentation model.

\subsection{Cross-Dataset Evaluation
\label{sec:evaluation}}

We train on one dataset and test on another dataset, where there are distribution shifts between the train and the test data. 
We propose two different evaluation approaches.

\paragraphbf{Matched Class Evaluation (Expt. 1).} The first approach creates two datasets that share the same classes, but have a significant disparity in the train and test data distribution. 
Classes that are common to the two initial datasets are identified following the procedure described in TruZe~\cite{truze}.
We describe the procedure and curated datasets in Supplementary Material.
This is the most realistic method to evaluate on a distribution shift, but it requires some manual procedures as well as finding datasets that share largely similar classes.

\paragraphbf{Cosine Similarity Evaluation (Expt. 2).} The second method applies the feature extractor trained on the source dataset to the videos in the target dataset, which are split into a training subset and a test subset.
This is a simpler method that does not require manual class matching. A more descriptive procedure can be found in the Supplementary Material.

Formally, the class prototype $\mathbf{c}_k \in \mathbb{R}^M$ for each class $k$ is the M-dimensional  mean vector of the embedded features belonging to that class in the training subset. 
Let $S_k$ be the set of videos in the training subset of the target dataset from class $k$.
$\mathbf{c}_k$ is computed by the following, where $y_i = k$ for all $ (\mathbf{x}_i, y_i) \in S_k$:
\begin{align}
\mathbf{c}_k = \frac{1}{|S_k|} \sum_{(\mathbf{x}_i, y_i) \in S_k}{h_\omega(\mathbf{x}_i)}
\end{align}
\noindent $h_\omega()$ is the embedded feature extraction function
computed by training on the source dataset. 

For a given test video from the test subset of the target dataset $\mathbf{x}_i \in S^{test}_k$, 
the probability of a given class label $k$ is estimated using the cosine similarity of the embedding to the target dataset's class prototype $\mathbf{c}_{k}$, followed by softmax. 
Given the cosine similarity function $d(\mathbf{x}, \mathbf{c}) = \frac{\mathbf{x} \cdot \mathbf{c}}{\norm{\mathbf{x}}\norm{\mathbf{c}}}$
we get:
\begin{align}
P(y=k|\mathbf{x}_i \in S^{test}_k) = \frac{\exp(-d(h_\omega(\mathbf{x}_i),\mathbf{c}_{k}))}{\sum_{k'}{\exp(-d(h_\omega(\mathbf{x}_i),\mathbf{c}_{k'}))}}
\end{align}
\noindent where $y$ denotes the class label.
If the largest estimated $P(\cdot)$ is for the same class as the ground truth, then this is a successful classification. Accuracy is computed over all samples in the test subset of the target dataset.

The parameter $\omega$ is not tuned during this operation.
That is, the target dataset does not contribute to fine-tuning the model. 
The motivation of this approach is to measure the transferability of the model from a source dataset to a target dataset without actually tuning the model parameters, showing the robustness of the model to sample distribution shift.

\section{Experiments}
\label{sec:experiments}

\paragraphbf{HMDB/Kinetics and UCF/Kinetics Datasets (Expt. 1).} HMDB-51~\cite{kuehne2011hmdb} is a popular human action recognition dataset that is composed of around 7,000 video clips divided into 51 categories, collected mainly from movies as well as YouTube. 
The UCF-101~\cite{ucf101} dataset consists of 13,320 videos in 101 action classes collected from YouTube. 
Kinetics-400~\cite{kinetics} is a large-scale action dataset in which each video clip is around 10 seconds long, and there are over 300,000 videos in 400 classes.

Training datasets are created from subsets of HMDB-51 or UCF-101 and the subset of the Kinetics-400 test set is used for testing.
The subsets share the same classes between the training and test sets.
The motivation for this approach to creating the datasets is to simulate a real-world environment where the test data comes from many unknown sources, with many variations in actions, capture conditions, aspect ratio, and so on. 
The Kinetics-400 dataset has many more samples in the fine-grained classes, so it was used for testing.

TruZe~\cite{truze} is used to identify shared classes from the HMDB-51 and Kinetics-400, and UCF-101 and Kinetics-400 datasets, based on the visual and semantic similarity. More details on this procedure can be found in the Supplementary Material.
The final datasets are named HMDB-28/Kinetics-28 and UCF-65/Kinetics-65, where the training sets are subsets of the HMDB and UCF data, respectively, and test subsets come from Kinetics. The 28 and 65 refer to the number of shared classes.
The three official published splits of HMDB-51 and UCF-101 are used, but only the shared classes are selected. The results in \Cref{tab:result} are the average performance over the three splits. 
The HMDB-28/Kinetics-28 dataset consists of 3445 HMDB and 43406 Kinetics videos, and the UCF-65/Kinetics-65 dataset consists of 8935 UCF and 78583 Kinetics videos.

\paragraphbf{HMDB $\leftrightarrow$ UCF Evaluation (Expt. 2).} For the experiments using the cosine similarity function, the HMDB-28 trained models are tested on UCF-101, and the UCF-65 trained models are tested on HMDB-51, using the cosine similarity measure with class prototypes as predictions. 
The  feature extractor 
trained using the training data is then used on the target dataset. 
It is used to create class prototype vectors (for each target class) from one part of the target dataset, and evaluation is based on the classification accuracy of the other part of the dataset. 
This assesses the quality of the feature extractor on another dataset with distribution shifts.
The results in \Cref{tab:result} are the average performance over the nine splits, three runs from the source dataset and each run evaluates with three splits from the target dataset.

See Supplementary Material for the resulting matching datasets in \Cref{tab:dataset_classes}, and a summary table of all experimental datasets (Expt. 1 and Expt. 2), in \Cref{tab:datasets}.

\paragraphbf{Augmentation Methods.} Results are compared for four different training approaches: training without augmentation, with random augmentation, with adversarial augmentation, and with curriculum AT~\cite{cai2018curriculum} as described in \Cref{sec:adversarial_method}. Experiments used the popular efficient 2D TSM model~\cite{tsm} with a ResNet50 backbone, Video Swin Transformer~\cite{swin} Tiny, and Uniformer-S~\cite{li2022uniformer} model. ImageNet pre-trained models were used instead of Kinetics pre-trained, so that the models never get to see the Kinetics data distribution.

Augmentation used translation to a maximum of 28 pixels, $10^\circ$ rotation, shear transform of 0.1, and scale from 0.9 to 1.5. 
For curriculum training, no augmentation was used for 20 epochs, then AT with a zero learning rate of the augmentation model was used for 20 more epochs. 
Then, AT with a triangular learning rate scheduling from 0.1 to 1.0 on the augmentation model was used for the rest of the training, except for the Uniformer model.
For Uniformer, the above was done for only 20 epochs, and then trained with random augmentation for 20 more epochs and fine-tuning with no augmentation for the rest to mitigate under-fitting issues.

See the Supplementary Material for implementation details.

\section{Results}

\begin{table}[t!]
\centering
\resizebox{0.495\linewidth}{!}{
\begin{tabular}{|l|l|l|c|c|}
\hline
\multirow{3}{*}{Model} & \multirow{3}{*}{Train Dataset} & \multirow{3}{*}{Augmentation} & \multicolumn{2}{c|}{Test Accuracy} \\
\cline{4-5}
 &&& Kinetics-65 & HMDB-51\\
&&& Matched Expt 1a & Cosine Expt 2a\\
\hline
\hline
\multirow{4}{*}{TSM} & \multirow{4}{*}{UCF-65 [77.50]} & None & 39.13 \textpm{} 0.56 & 37.61 \textpm{} 1.88 \\ 
&& Random & 40.90 \textpm{} 0.32 & 38.19 \textpm{} 1.39 \\ 
&& Adversarial & 42.42 \textpm{} 0.63 & 38.99 \textpm{} 1.26 \\ 
&& Curriculum & \textbf{42.51 \textpm{} 0.62} & \textbf{39.14 \textpm{} 1.21} \\ 
\hline
\multirow{4}{*}{Swin} & \multirow{4}{*}{UCF-65 [81.15]} & None & 37.08 \textpm{} 1.43 & 38.91 \textpm{} 1.63 \\ 
&& Random & 40.80 \textpm{} 1.90 & 40.61 \textpm{} 1.40 \\ 
&& Adversarial & \textbf{42.27 \textpm{} 0.24} & 41.48 \textpm{} 1.58 \\ 
&& Curriculum & 41.20 \textpm{} 0.72 & \textbf{41.58 \textpm{} 1.63} \\ 
\hline
\multirow{4}{*}{Uniformer} & \multirow{4}{*}{UCF-65 [52.05]} & None & 18.78 \textpm{} 0.22 & 21.39 \textpm{} 1.32 \\ 
&& Random & 22.42 \textpm{} 0.98 & 24.92 \textpm{} 1.66 \\ 
&& Adversarial & 22.95 \textpm{} 0.68 & \textbf{26.16 \textpm{} 2.10} \\ 
&& Curriculum & \textbf{23.61 \textpm{} 0.27} & 24.93 \textpm{} 1.60 \\ 
\hline
\end{tabular}
}
\resizebox{0.495\linewidth}{!}{
\begin{tabular}{|l|l|l|c|c|}
\hline
\multirow{3}{*}{Model} & \multirow{3}{*}{Train Dataset} & \multirow{3}{*}{Augmentation} & \multicolumn{2}{c|}{Test Accuracy} \\ \cline{4-5} &&& Kinetics-28 & UCF-101\\
&&& Matched Expt 1b & Cosine Expt 2b\\
\hline
\hline
\multirow{4}{*}{TSM} & \multirow{4}{*}{HMDB-28 [55.45]} & None & 29.75 \textpm{} 1.08 & 60.51 \textpm{} 0.79\\ 
&& Random & 30.13 \textpm{} 0.42 & 61.21 \textpm{} 0.57 \\ 
&& Adversarial & 32.44 \textpm{} 0.48 & 62.60 \textpm{} 0.59 \\ 
&& Curriculum & \textbf{32.82 \textpm{} 1.41} & \textbf{63.18 \textpm{} 0.71} \\ 
\hline
\multirow{4}{*}{Swin} & \multirow{4}{*}{HMDB-28 [54.67]} & None & 25.26 \textpm{} 0.80 & 59.88 \textpm{} 0.55 \\ 
&& Random & 26.63 \textpm{} 0.97 & 60.99 \textpm{} 1.30 \\ 
&& Adversarial & 27.31 \textpm{} 0.57 & 62.57 \textpm{} 0.72 \\ 
&& Curriculum & \textbf{27.60 \textpm{} 0.62} & \textbf{62.95 \textpm{} 0.82} \\ 
\hline
\multirow{4}{*}{Uniformer} & \multirow{4}{*}{HMDB-28 [29.43]} & None & 14.53 \textpm{} 0.51 & 28.23 \textpm{} 0.94 \\ 
&& Random & 14.33 \textpm{} 1.03 & 28.67 \textpm{} 1.36 \\ 
&& Adversarial & 15.13 \textpm{} 0.79 & 29.88 \textpm{} 2.62 \\ 
&& Curriculum & \textbf{15.25 \textpm{} 0.75} & \textbf{30.83 \textpm{} 1.04} \\ 
\hline
\end{tabular}
}
\vspace{1pt}
\caption{Results from three models (TSM ResNet50, Video Swin Transformer Tiny, and Uniformer-S), two training datasets (UCF-65 and HMDB-28), four augmentation strategies (no augmentation, random, adversarial, and curriculum), and two test datasets (Kinetics, and HMDB/UCF). In all cases, adversarial augmentation or curriculum adversarial augmentation training outperformed all baselines.
The columns labeled Test Accuracy show the performance on the target test set. The values in brackets in the Train Dataset columns show the ``no augmentation'' accuracy on the test set of the same dataset to demonstrate the performance drop when evaluating to the Kinetics dataset.
}
\label{tab:result}
\vspace{-20pt}
\end{table}

\subsection{Shared Class Experiments 1a, 1b}
When the adversarial training approaches presented in \Cref{sec:adversarial_method} are applied, target dataset performance improves.
\Cref{tab:result} summarizes the main cross-dataset evaluation results for the four augmentation strategies presented in \Cref{sec:adversarial_method}. The different adversarial augmentation strategies gave improved accuracy for the target datasets (see Kinetics-28 and 65 results columns).

Also, unlike what is reported in \cite{Schiappa_2023_CVPR}, the convolutional architecture performed better with the distribution shift compared to transformer models in most of the cases except for the Swin Transformer trained on UCF-65 and tested on HMDB-51. 
We hypothesize that this is because the Kinetics test dataset shows natural and realistic distribution shifts. 
In addition, we did not use the Kinetics pre-trained models, and the transformer architectures require large-scale data to reach the maximum potential.

In all cases, the adversarial augmentation or curriculum methods outperform all baselines, given a fixed network architecture, for all training and test datasets. 
Although the ``random augmentation'' and ``adversarial augmentation'' allow an identical range of transforms, generating adversarial examples through gradient ascent produces ``harder than random'' augmentation which improves the overall performance. Furthermore, adding the simple curriculum mostly improved over the adversarial benchmark. 

See the Supplementary Material for confusion matrices that show per-class performance drop with distribution shifts and improvements using the proposed adversarial augmentation.

\subsection{Cosine Similarity Evaluation 2a, 2b}

The cosine-similarity-based accuracy results on HMDB-51 and UCF-101 are shown in the Cosine Expt columns of \Cref{tab:result}.
The results follow a very similar trend to the ``realistic'' Kinetics Expt 1. 
The advantage of using this accuracy measure as compared to testing on Kinetics with overlapping classes is that it requires no thorough analysis of the source and target datasets to find overlapping classes, making it simple to set up the cross-dataset experiments even using new datasets.

\section{Conclusion}
This paper addressed the problem of model generalization to realistic test distribution shifts. 
Two new datasets that are comprised of three existing datasets were created that share the same subset of label classes. 
Although the same classes were used, the variety of videos in the original dataset sources meant that there was a huge distribution shift from the source to the target datasets.
When using the target datasets, action recognition performance dropped significantly.
This led to trying adversarial augmentation, with and without curriculum scheduling, as an approach to generating hard adversarially augmented videos. 
This approach gave a small but meaningful improvement in performance, even with the large distribution shift in the test data.
The second cross-dataset evaluation approach, using the cosine similarity as logits, also showed a similar trend as the matching dataset experiments, providing a simpler alternative method without having to curate datasets with matching classes.

\pagebreak

\appendix

{\noindent\Large\textbf{Supplementary Material}}

\section{TruZe Class Matching Procedure}

For Expt. 1a and 1b, we follow TruZe~\cite{truze} to identify overlapping classes in two different datasets. 
To choose the common classes, visual features are extracted on two existing datasets using the I3D~\cite{i3d} model pre-trained on Kinetics-400~\cite{kinetics} and 
semantic cues are extracted using the sen2vec~\cite{sen2vec} model pre-trained on Wikipedia.
The visual and semantic similarity features are combined and then used in the TruZe~\cite{truze} procedure (\ie include exact matches, matches that can be either superset or subset, and matches that predict the same visual and semantic matches) to obtain a set of extremely similar classes from the two source action recognition datasets. 
The matched classes from the two datasets are verified manually and a subset is selected that has a substantial overlap in visual and semantic cues. One dataset is used for training and validation, and the other dataset is used for testing.

In TruZe, normally, classes from UCF or HMDB that have overlapping context with the corresponding Kinetics class are removed, so that using the Kinetics pre-trained models would not bias the zero-shot settings. 
However, in our robustness problem, the train and test datasets are created with the \emph{opposite} goal, where only overlapping classes are selected. 
For instance, the class ``climb'' in HMDB-51 is treated as the same class as ``rock climbing'', ``ice climbing'', ``climbing a rope'', and ``climbing tree'' in Kinetics-400. Examples of the resulting splits for the matched class experiments (Expt 1a and 1b) can be found in \Cref{tab:dataset_classes}.

\section{Cosine Similarity Evaluation Procedure}
In Expt 2a and 2b, we use cosine similarity between class prototypes and features on the evaluating dataset to estimate predictions. We then report accuracy with the simulated predictions to compare the performance of augmentation strategies given distribution shifts.

We first train the classification model using the source dataset. Since the number and types of classes of the training (source) and evaluation (target) datasets are different, we detach the last classification layer of the source dataset classification model to use it as a feature extractor. We then calculate class prototypes of all the classes in the target dataset by simply averaging the features of each class in the training set, inspired by~\cite{snell2017prototypical}.
The prediction score of a sample from the target dataset's test set is estimated using the cosine similarity of their feature vector and the class prototypes, followed by softmax.
In other words, the cosine similarity is used as logits, which are formed by producing high activations on classes that are closely aligned to the training set of the target dataset.
Using these estimated prediction probabilities, we report accuracy.
We will show that adversarial augmentation of the source dataset also produces improved classification of the target dataset, by producing a more robust feature extractor for use with this similarity measure.
Note that the source dataset is never used for evaluation, and only be used for training the model, and the target dataset's training set does not contribute to fine-tuning the model.
The role of all four splits in two different datasets can be found in \Cref{tab:cosine_procedure}.

\section{Datasets}

\Cref{tab:datasets} summarizes all datasets used for the experiments (Expt. 1 and Expt. 2).

\begin{table*}[t]

\centering
\resizebox{\linewidth}{!}{
\begin{tabular}{|l|l|c|c|}
\hline
HMDB-51 Classes & Kinetics-400 Classes \\
\hline
brush hair & curling hair, dying hair, brushing hair \\
cartwheel & cartwheeling, somersaulting \\
catch, throw & shooting goal, juggling, catching or throwing frisbee, catching or throwing baseball, catching or throwing softball, throwing axe, throwing ball, throwing discus \\
clap & clapping, applauding \\
climb & rock climbing, ice climbing, climbing tree, climbing a rope \\
dive & diving cliff, bungee jumping \\
dribble & dribbling basketball \\
drink & drinking beer, drinking shots, drinking \\
eat	& eating burger, eating cake, eating carrots, eating chips, eating doughnuts, eating hotdog, eating ice cream, eating spaghetti, eating watermelon \\
golf & golf driving, golf chipping, golf putting \\
$\cdots$ & $\cdots$ \\
\hline
\multicolumn{2}{c}{}\\[-0.1em]
\hline
UCF-101 Classes & Kinetics-400 Classes \\
\hline
Basketball & shooting basketball, dribbling basketball, playing basketball \\
BasketballDunk & dunking basketball \\
BodyWeightSquats & lunge, squat, dead lifting \\
BreastStroke & swimming breast stroke, swimming back stroke \\
CleanAndJerk & clean and jerk, snatch weight lifting \\
CliffDiving & diving cliff, springboard diving \\
Haircut	& shaving head, braiding hair, getting a haircut \\
HorseRiding & riding or walking with horse, riding mule \\
PlayingPiano & playing piano, playing organ \\
Skiing & skiing (not slalom or crosscountry), skiing crosscountry, skiing slalom \\
$\cdots$ & $\cdots$ \\
\hline

\hline
\end{tabular}
}
\caption{The HMDB-28/Kinetics-28 (above) and UCF-65/Kinetics-65 (below) datasets which are subsets from the original HMDB-51 UCF-101, Kinetics-400 datasets. Visually and semantically similar classes were combined which allows testing with real-life data from YouTube that has a significant distribution shift from the training data. The HMDB or UCF data are used for training and validation, and the Kinetics data are used for testing.
\label{tab:dataset_classes}}
\end{table*}

\begin{table}[t]

\centering
\resizebox{0.99\linewidth}{!}{
\begin{tabular}{|l|l|l|l|}
\hline
Dataset Type & \# Classes & Split & Usage \\
\hline
\hline
\multirow{2}{*}{Source} & \multirow{2}{*}{$N$} & train & Training \\
&& test & Validation \\
\hline
\multirow{2}{*}{Target} & \multirow{2}{*}{$K$} & train & Obtaining $K$ class prototypes \\
&& test & Testing with cosine similarity as logits \\
\hline
\end{tabular}
}
\caption{Purpose of all dataset splits for the cosine similarity evaluation method (Expt. 2).
Given the four splits of the data, the source dataset is used to train (with or without AT) and validate the model. The target dataset is used to evaluate the model without further fine-tuning.
}
\label{tab:cosine_procedure}
\end{table}

\begin{table}[t!]

\centering
\resizebox{0.99\linewidth}{!}{
\begin{tabular}{|c|l|l|l|}
\hline
Expt & Name & Original Dataset & Usage \\
\hline
\hline
\multirow{2}{*}{1a} & UCF-65 & UCF-101 & Training, validation \\
& Kinetics-65 & Kinetics-400 & Testing UCF-65 trained models \\
\hline
\multirow{2}{*}{1b} & HMDB-28 & HMDB-51 & Training, validation \\
& Kinetics-28 & Kinetics-400 & Testing HMDB-28 trained models \\
\hline
\multirow{2}{*}{2a} & UCF-65 & UCF-101 & Training, validation \\
& HMDB-51 & (original) & Testing UCF-65 trained models w/ cosine similarity \\
\hline
\multirow{2}{*}{2b} & HMDB-28 & HMDB-51 & Training, validation \\
& UCF-101 & (original) & Testing HMDB-28 trained models w/ cosine similarity \\ 
\hline
\end{tabular}
}
\caption{List of datasets used for our experiments. The HMDB-51, UCF-101, and Kinetics-400 are the original datasets. The UCF-65, Kinetics-65, HMDB-28, and Kinetics-28 are the proposed subsets.
}
\label{tab:datasets}
\end{table}

\section{Implementation Details.} Videos were resized to 224 $\times$ 224 and sampled to 8 frames sparsely similar to \cite{tsn}.
The classifier models were trained for 200 epochs using an SGD optimizer with an initial learning rate of 0.0001, decaying the learning rate using cosine annealing scheduling. 
For adversarial training (AT), the augmentation model used a learning rate of 0.1. 
For adversarial plus curriculum training, the $\mathcal{L_{\text{AT-ME}}}$ loss was used with $\alpha=0.5$ and $\gamma = 0.5$. 
Two NVIDIA RTX 3090 GPUs were used with batch size 16 to train the TSM models, an NVIDIA A100 GPU with batch size 16 and 32 for training the Uniformer and Swin Transformer models, respectively. 

\section{Adversarial Augmentation Examples}

\begin{figure}[t]
    \centering
    \includegraphics[width=0.95\columnwidth]{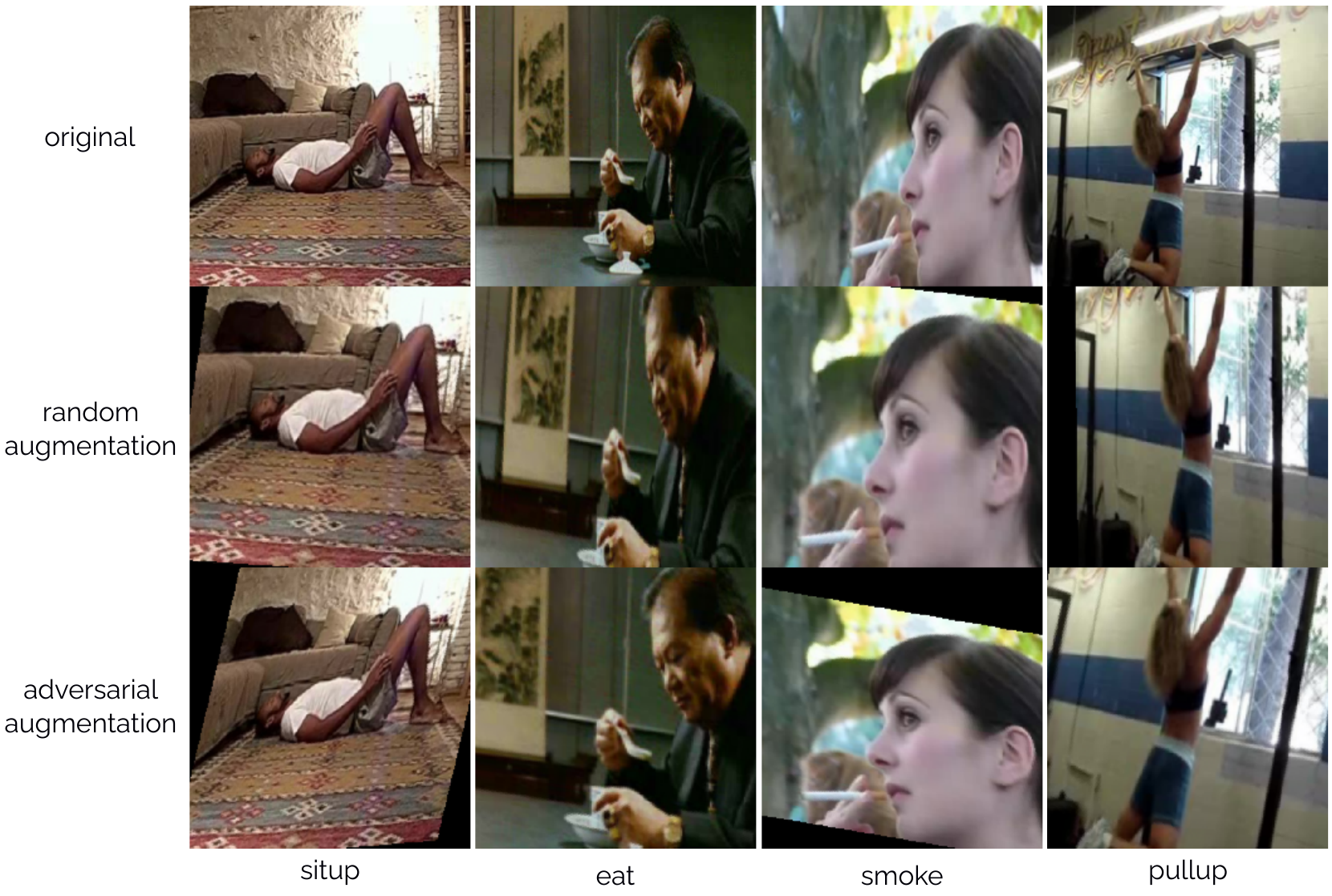}
    \caption{Examples of the vanilla and proposed augmentation examples in the HMDB-51 dataset while training the Video Swin Transformer model.}
    \label{fig:augmentation}
\end{figure}

Some examples of adversarial augmentations in comparison to random augmentations are depicted in \Cref{fig:augmentation}. One might think that the most extreme and unrealistic augmentations will be challenging to the classifier. However, adversarial augmentation can sometimes render more realistic yet challenging examples.

\section{Performance Drop with Distribution Shifts}

\begin{figure}[t!]
    \centering
    \begin{subfigure}[t]{\columnwidth}
        \centering
        \includegraphics[trim={15cm 5cm 12cm 4cm},clip,width=0.6\columnwidth]{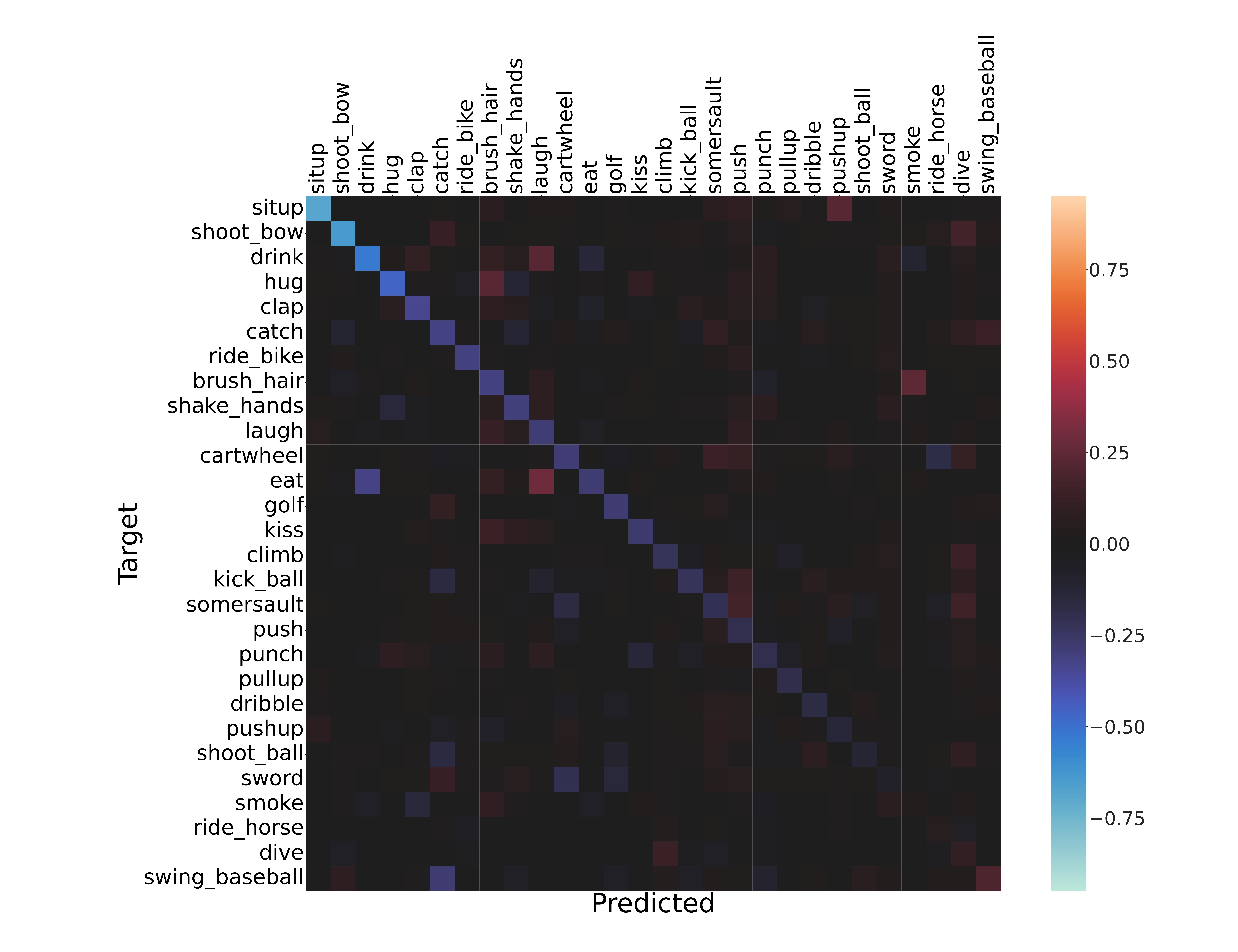}
        \caption{HMDB-28}
        \label{fig:confusion_hmdb}
    \end{subfigure}
    \begin{subfigure}[t]{\columnwidth}
        \centering
        \includegraphics[trim={17cm 5cm 11cm 4cm},clip,width=0.6\columnwidth]{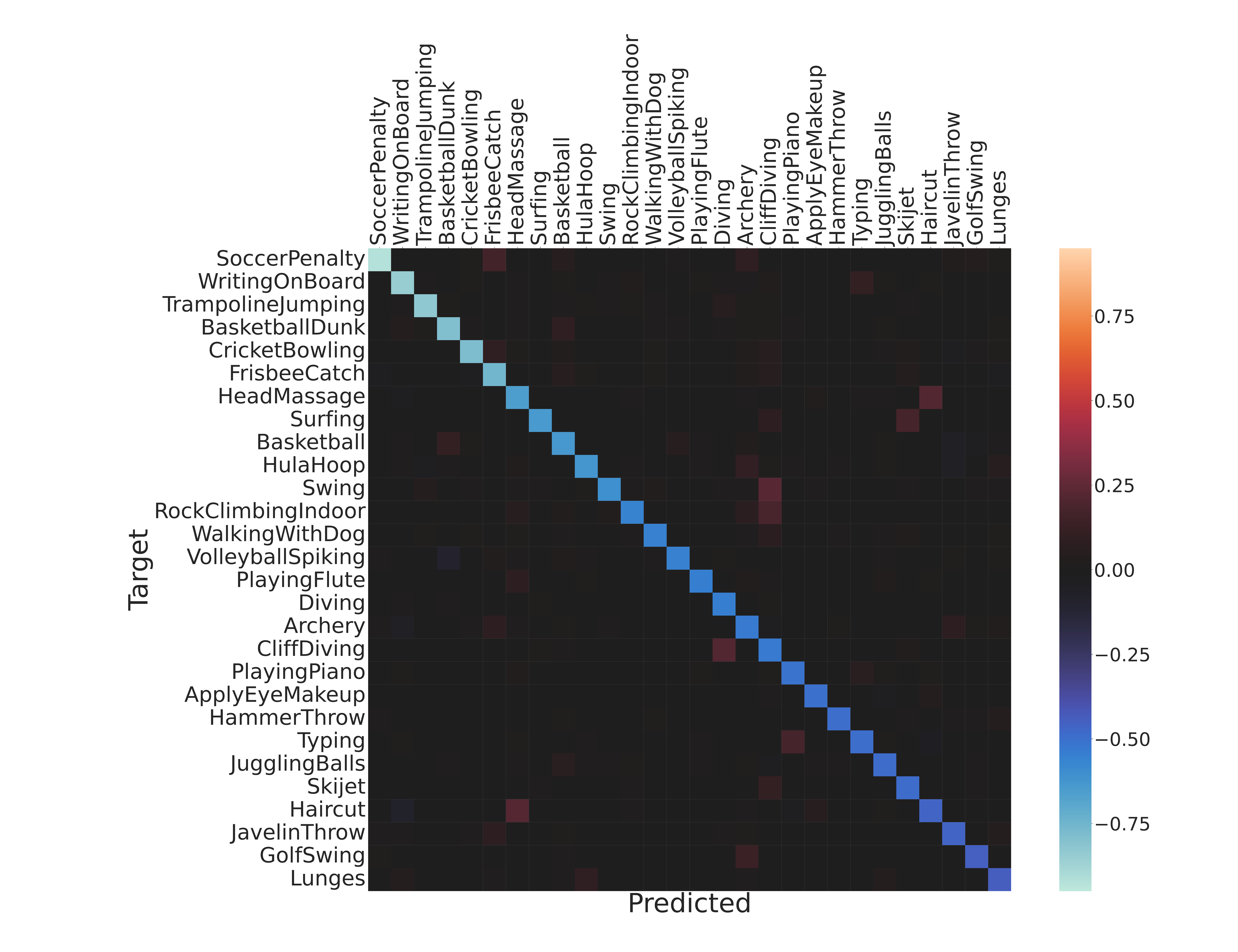}
        \caption{UCF-65 (top 28 classes with the highest accuracy drop)}
        \label{fig:confusion_ucf}
    \end{subfigure}

    \caption{Confusion matrix difference between evaluating on the same dataset (a:HMDB at top or b:UCF at bottom) and a different one (Kinetics), given a TSM model with no augmentation strategy. The negative (blue) values on the diagonal line indicate the class accuracy drop when evaluated on Kinetics.
    Almost every class has a drastic drop.
    Overall, there is a 25.7\% and 38.37\% accuracy drop for the HMDB and UCF training datasets, respectively.
    \label{fig:confusion_matrix_diff}}
    \vspace{-10pt}
\end{figure}

Two confusion matrices are created collating the performance when training on HMDB-28 and testing on either HMDB-28 or Kinetics-28 (test sets) similar to Expt 1a and 1b.
The former is subtracted from the latter to record how much difference there is in the performance, which is shown in
\Cref{fig:confusion_matrix_diff}(a). 
(The figure also shows the same results when using UCF-65 and Kinetics-65.) 
In this figure, no augmentation strategy is used.
The figure shows that 
there is a dramatic performance drop when the models are tested on a different dataset (Kinetics) to training (HMDB or UCF).
Some categories such as ``situp'', ``shoot\_bow'',
and ``drink'' are damaged more severely. 
As seen in \Cref{fig:drink}, HMDB/UCF videos have objects and actors that are clearly visible and stable in the frame, while Kinetics videos tend to have lots of camera motion with different sizes of the objects.
The overall accuracy drop from UCF-65 to Kinetics-65 is even more significant. 
These results show that the raw trained model is not robust to distribution shifts between datasets, which is not ideal for deployment in real applications. %

\section{Improvements on Classes with Larger Distribution Shifts}
\begin{figure}[t!]
    \centering
    \begin{subfigure}[t]{\columnwidth}
        \centering
        \includegraphics[trim={15cm 5cm 12cm 4cm},clip,width=0.6\columnwidth]{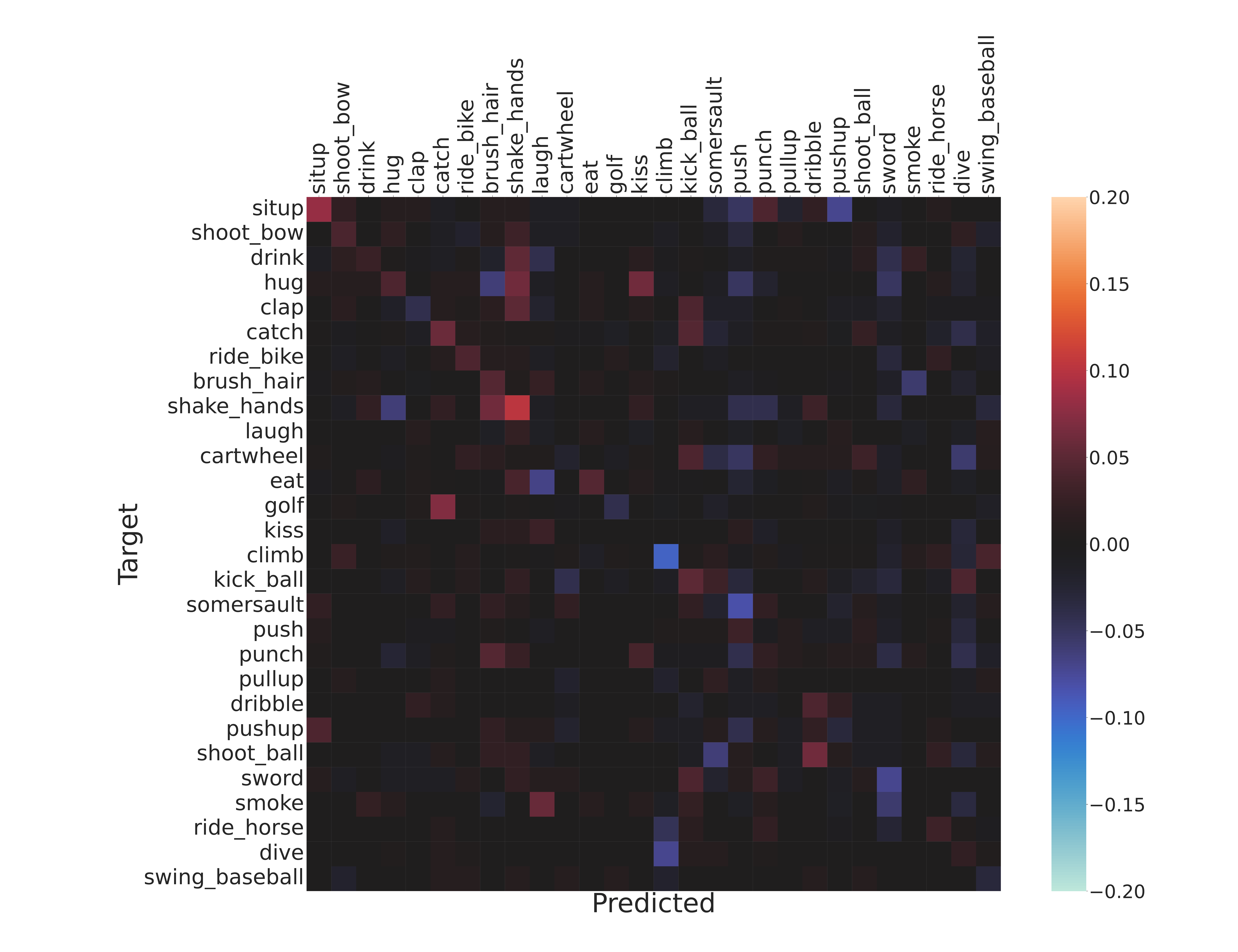}
        \caption{HMDB-28}
        \label{fig:confusion_improv_hmdb}
    \end{subfigure}
    
    \begin{subfigure}[t]{\columnwidth}
        \centering
        \includegraphics[trim={17cm 5cm 11cm 4cm},clip,width=0.6\columnwidth]{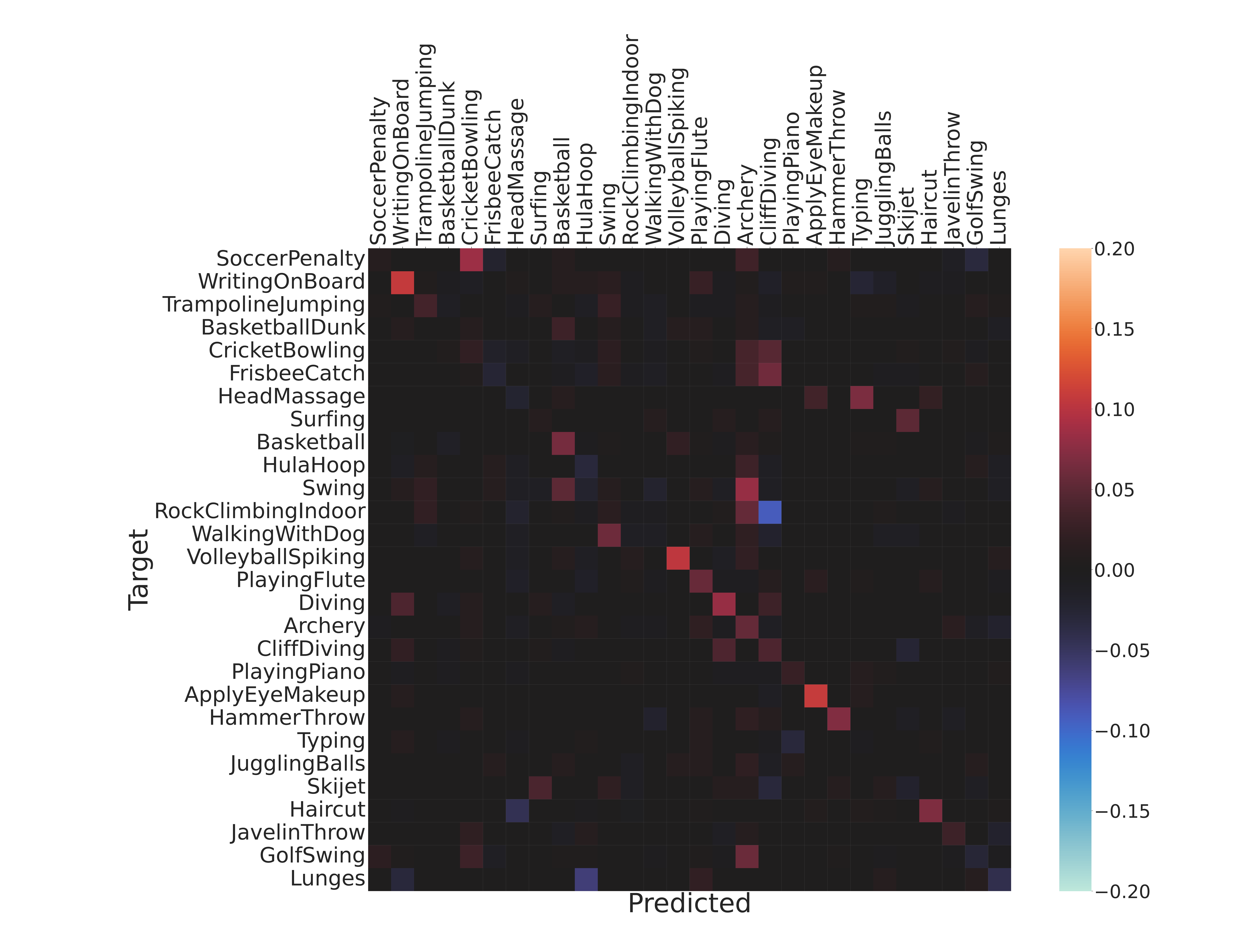}
        \caption{UCF-65 (top 28 classes with the highest accuracy drop)}
        \label{fig:confusion_improv_ucf}
    \end{subfigure}

    \caption{Confusion matrix difference between no augmentation strategy and the proposed curriculum adversarial augmentation training with the TSM model and evaluation on Kinetics. The positive (red) values on the diagonal line indicate the added class-wise accuracy by using the proposed approach. The classes are sorted by the drop in performance using the proposed cross-dataset evaluation, as seen in \Cref{fig:confusion_matrix_diff}.
    }
    \vspace{-10pt}
    \label{fig:confusion_matrix_improvements}
\end{figure}

To demonstrate the improvement in transfer performance by the use of the curriculum adversarial strategy, we computed the difference of the confusion matrices for None and Curriculum augmentation in Expt. 1a and 1b.
\Cref{fig:confusion_matrix_improvements} shows 
the results, where the reddish boxes on the diagonal indicate improved performance. It is clear that the proposed adversarial augmentation makes the model more robust on the classes with the larger distribution shifts, as presented in \Cref{fig:confusion_matrix_diff}.
This supports our claim that the proposed adversarial augmentation makes the model more robust in classes with a
huge distribution shift.

\end{document}